%% file: main.tex
\documentclass[10pt,twocolumn,letterpaper]{article}

\usepackage[pagenumbers]{cvpr} 
\usepackage{multirow}
\usepackage{xcolor}
\usepackage{tcolorbox}
\usepackage{colortbl}
\input{preamble}
\definecolor{cvprblue}{rgb}{0.21,0.49,0.74}
\usepackage[pagebackref,breaklinks,colorlinks,allcolors=cvprblue]{hyperref}

\def\paperID{4} 
\def\confName{CVPR}
\def\confYear{2026}

\title{Unlocking the Potential of Grounding DINO in Videos: Parameter-Efficient Adaptation for Limited-Data Spatial-Temporal Localization}

\author{
\textbf{Zanyi Wang$^{1}$\thanks{Equal contribution. This work was completed during the internship at SGIT AI Lab.}} \quad
\textbf{Fan Li$^{1*}$} \quad
\textbf{Dengyang Jiang$^{1}$} \quad
\textbf{Liuzhuozheng Li$^{1}$} \quad
\textbf{Yunhua Zhong$^{2}$} \\
\textbf{Guang Dai$^1$} \quad
\textbf{Mengmeng Wang$^{3,1}$\thanks{Corresponding author.}} \\[2mm]
{\small 
$^1$SGIT AI Lab, State Grid Corporation of China \quad
$^2$University of HongKong \quad
$^3$Zhejiang University of Technology
}
}

\begin{document}
\maketitle
\input{sec/0_abstract}    
\input{sec/1_intro}

\input{sec/2_method}
\input{sec/3_experiment}

\input{sec/4_conclusion}

{
    \small
    \bibliographystyle{ieeenat_fullname}
    \bibliography{main}
}


\end{document}

%% file: sec/0_abstract.tex
\begin{abstract}
Spatio-temporal video grounding (STVG) aims to localize queried objects within dynamic video segments. Prevailing fully-trained approaches are notoriously data-hungry. However, gathering large-scale STVG data is exceptionally challenging: dense frame-level bounding boxes and complex temporal language alignments are prohibitively expensive to annotate, especially for specialized video domains. Consequently, conventional models suffer from severe overfitting on these inherently limited datasets, while zero-shot foundational models lack the task-specific temporal awareness needed for precise localization.
To resolve this small-data challenge, we introduce ST-GD, a data-efficient framework that adapts pre-trained 2D visual-language models (e.g., Grounding DINO) to video tasks. To avoid destroying pre-trained priors on small datasets, ST-GD keeps the base model frozen and strategically injects lightweight adapters (~10M trainable parameters) to instill spatio-temporal awareness, alongside a novel temporal decoder for boundary prediction. This design naturally counters data scarcity. Consequently, ST-GD excels in data-scarce scenarios, achieving highly competitive performance on the limited-scale HC-STVG v1/v2 benchmarks, while maintaining robust generalization on the VidSTG dataset. This validates ST-GD as a powerful paradigm for complex video understanding under strict small-data constraints.

\end{abstract}

%% file: sec/1_intro.tex
\section{Introduction}
 \begin{figure}[htb]
  \centering
   
  \begin{subfigure}[b]{0.47\textwidth}
    \includegraphics[width=\textwidth]{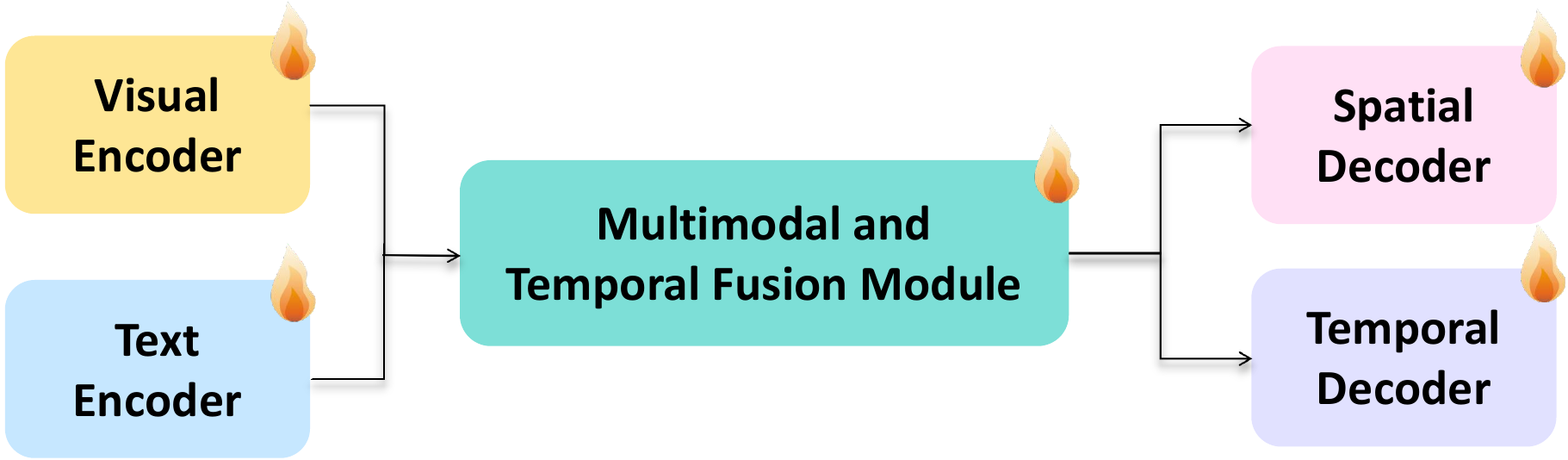} 
    \caption{Resources Consuming  due to 
Fully Trainable Modules  
}
    \label{fig:previous}
  \end{subfigure}
  \hfill 
  \begin{subfigure}[b]{0.47\textwidth}
    \includegraphics[width=\textwidth]{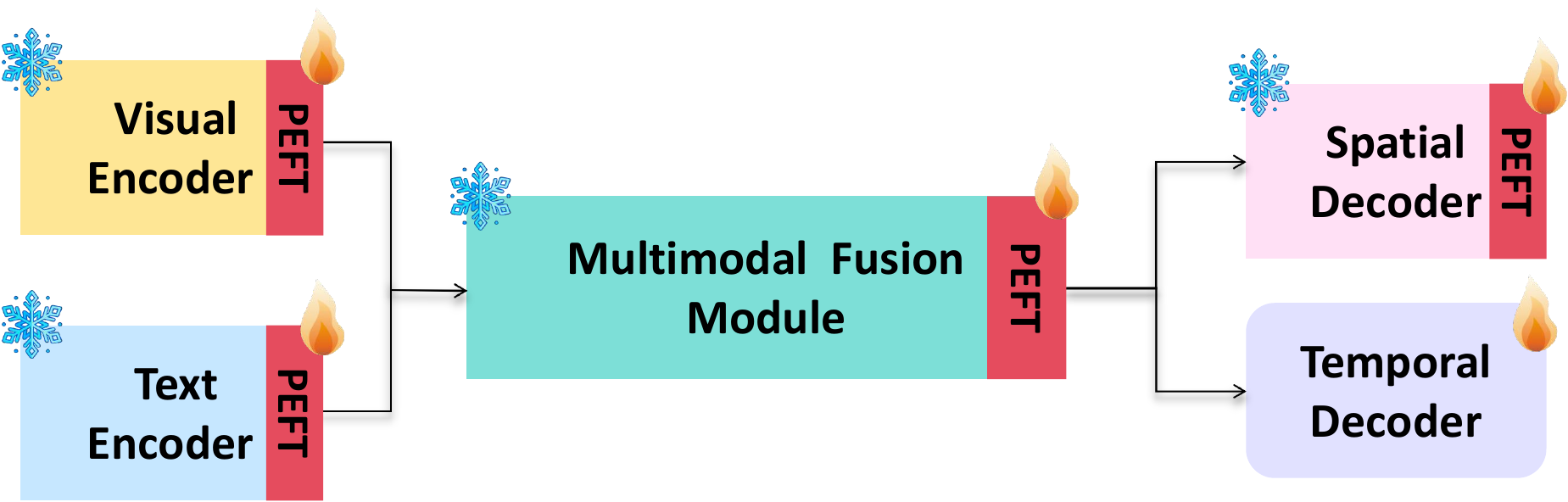} 
    \caption{Resources Friendly  by 
Efficient PEFT-Based Transfer
}
    \label{fig:our}
  \end{subfigure}

\caption{\textbf{(a)} Existing fully-trainable STVG methods are data-hungry and overfit on limited datasets. \textbf{(b)} Our ST-GD adapts Grounding DINO via PEFT. Training only lightweight adapters preserves pre-trained priors, enabling highly data-efficient transferring learning.}
  
   \vspace{-1em}
  \label{fig:main_figure}
\end{figure}

Spatio-temporal video grounding (STVG) \cite{zhang2020does,yang2022tubedetr} is a task that requires localizing an object described by a natural language query in both space and time. This task is crucial for enabling machines to understand videos and powers applications like autonomous driving~\cite{auto1,auto2} and interactive video editing~\cite{ved1,ved2}. However, STVG is inherently challenging due to the complexities of modeling spatio-temporal dependencies and bridging the language-vision semantic gap.

\par Previous methods have made some attempts to solve these problems. 2D-TAN \cite{tan2021augmented} tackles STVG task by first temporally grounding the relevant moment using an augmented 2D-TAN and then identifying the human within that segment based on designed rules. Tubedetr \cite{yang2022tubedetr} addresses this task by employing a transformer-based architecture with video-text encoder that models spatial multimodal interactions over frames, and a space-time decoder that jointly performs spatial-temporal localization. STCAT \cite{jin2022embracing} and CGSTVG \cite{gu2024context} follow Tubedetr style but add designed temporal modules to consider consistency and context to further improve performance.
Acquiring dense frame-level bounding boxes and precise temporal language alignments for videos is exceptionally challenging. Consequently, such a fully trainable approach is severely data-hungry; it struggles to learn generalizable representations and easily overfits the limited annotations available in typical STVG datasets \cite{wang2022learning}.

\begin{figure}[t]
  \centering
  \includegraphics[width=0.87\linewidth]{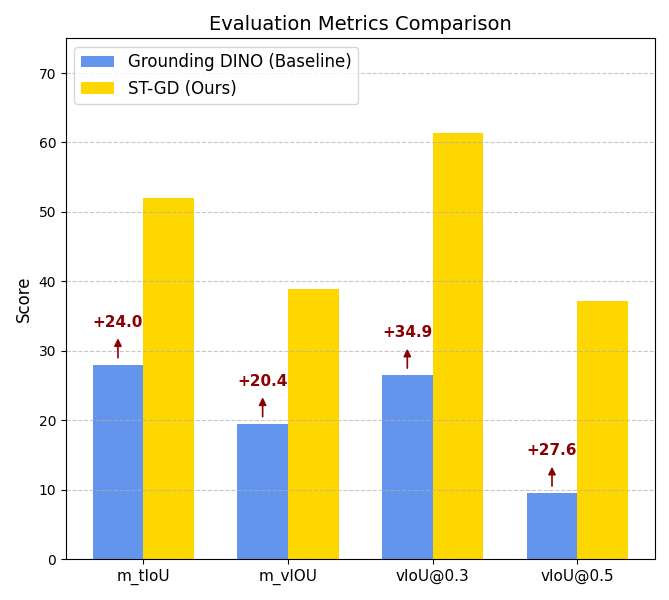}
  \vspace{-1.2em}
  \caption{Performance comparison on HCSTVG-v1 against a ``minimal baseline" (frozen Grounding DINO with fine-tuned heads) highlights naive adaptation's ineffectiveness and validates our specialized PEFT approach.
 }
 \vspace{-1.5em}
  \label{fig:compare}
\end{figure}

Recently, the remarkable success of foundation models like Grounding DINO \cite{liu2024grounding} on image-level tasks has inspired their adaptation for video. This has given rise to two dominant paradigms, each with its own trade-offs. The first involves extensive fine-tuning of these models on video data \cite{wasim2024videogrounding}. While this can adapt the model to the target domain, full fine-tuning on small-scale STVG datasets easily induces severe overfitting. It risks catastrophic forgetting of the invaluable, general-purpose knowledge learned during large-scale pre-training \cite{luo2025empirical, liu2025continuallearningvlmssurvey}, while still adding computational overhead. The second paradigm, often employed with even larger VLMs such as LLaVA-OneVision-7B \cite{lin2024video} or Qwen2-VL-7B \cite{wang2024qwen2}, relies on zero-shot or few-shot inference \cite{bao2024e3m}. This approach maximally preserves pre-trained knowledge but suffers from suboptimal adaptation, as the frozen model cannot be sufficiently tailored to the nuances of spatio-temporal localization. This leaves a critical question unanswered: How can we deeply adapt a powerful foundation model for STVG while retaining its robust pre-trained knowledge, all with minimal computational cost?

To address this challenge, we chart a third course: Parameter-Efficient Fine-Tuning (PEFT), as illustrated in Figure \ref{fig:our}. We propose ST-GD, a framework that, to the best of our knowledge, is the first to introduce PEFT to the STVG task. Instead of choosing between expensive fine-tuning and rigid zero-shot inference, ST-GD enables deep yet efficient adaptation. To validate our approach, we first demonstrate in Figure \ref{fig:compare} that a minimal baseline—fine-tuning only spatial and temporal heads on a frozen Grounding DINO—yields poor performance, confirming that simplistic adaptation is insufficient. In contrast, our framework strategically injects a series of lightweight, specialized adapters to instill spatio-temporal awareness. Furthermore, it introduces a novel temporal decoder that uniquely repurposes the base model’s internal, language-guided representations for temporal boundary prediction.

Finally, we conduct comprehensive experiments to evaluate the effect of ST-GD. Without bells and whistles, our approach achieves competitive performance on HCSTVG v1 and v2 datasets, while using approximately 4\% of those required by existing SOTA models, which once again demonstrates the efficiency and effectiveness of our method.

In summary, our main contributions are as follows:

\begin{itemize}
    \item \textbf{A Novel Paradigm for STVG Adaptation:} We pioneer the use of PEFT for STVG, proposing a general framework (ST-GD) that enables deep, efficient adaptation of static foundation models, charting a new path beyond traditional fine-tuning and zero-shot approaches.
    \item \textbf{Specialized Temporal Modeling via PEFT:} We design a set of lightweight, specialized adapters and a novel temporal decoder that uniquely leverages the base model's internal queries, effectively teaching temporal reasoning with minimal parameter updates.
    \item \textbf{Superior Performance under Data Constraints:} Our method achieves competitive results on challenging benchmarks with only a fraction of the trainable parameters (approx. 4\% of SOTA), establishing a robust and data-efficient standard for video grounding under limited-data regimes.
\end{itemize}

\vspace{1mm}
\noindent\textbf{Small Data Statement.}
(1) \textit{Why this research qualifies as small data research:} The task of Spatio-Temporal Video Grounding (STVG) intrinsically suffers from a severe lack of large-scale annotated data due to the prohibitive cost of dense, frame-level bounding box annotations and complex temporal-language alignment. Consequently, datasets in this domain (e.g., HC-STVG) are orders of magnitude smaller than typical image or text pre-training corpora.
(2) \textit{Methods employed to tackle this challenge:} To prevent catastrophic forgetting and severe overfitting on these limited video datasets, we propose a data-efficient PEFT framework, ST-GD. By keeping the large-scale foundational model frozen and uniquely integrating lightweight adapters, our method circumvents the data-hungry nature of video understanding, successfully transferring robust static priors to dynamic tasks using constrained training data.

\section{Related Work}

\subsection{Spatio-Temporal Video Grounding}
Spatio-temporal video grounding (STVG) localizes objects in untrimmed videos based on query sentences \cite{weng2024longvlm,zhang2020does}. Early methods like LOCVTP \cite{cao2022locvtp,tang2021human} used two stages: proposal generation and selection. Subsequent one-stage frameworks emerged, such as STVGBert \cite{su2021stvgbert} for simultaneous spatial and temporal grounding. TubeDETR \cite{yang2022tubedetr} improved multimodal fusion with transformers, while STCAT \cite{jin2022embracing} modeled spatio-temporal interactions end-to-end. CGSTVG \cite{gu2024context} enhanced temporal features with a 3D backbone. While innovative, this ``from-scratch" paradigm fundamentally limits their data efficiency and generalization capability, as they cannot inherit the rich, open-vocabulary understanding of vision-language foundation models pre-trained on large-scale data.

\subsection{Adapting Pre-trained Models for Grounding}
The success of large multimodal models like Qwen2-VL \cite{wang2024qwen2} Grounding DINO \cite{liu2024grounding} has spurred diverse adaptation strategies for downstream tasks. These range from continual pre-training on specialized data for domains like GUI Grounding \cite{cheng2024seeclick, you2024ferret}, to extensive fine-tuning for video tasks \cite{wasim2024videogrounding}, or even zero-shot inference with massive VLMs \cite{bao2024e3m}. However, these approaches create a difficult trade-off: fine-tuning is computationally expensive and risks catastrophic forgetting \cite{liu2025continuallearningvlmssurvey}, while zero-shot methods lack the deep, task-specific adaptation required for nuanced tasks like STVG.

\subsection{Parameter-Efficient Fine-Tuning for Vision}
Parameter-Efficient Fine-Tuning (PEFT) techniques, such as LoRA \cite{hu2022lora} and Adapters \cite{houlsby2019parameterefficienttransferlearningnlp}, have become standard for adapting Large Language Models. Their application in vision is also growing. Notably, ST-Adapter \cite{pan2022st} successfully applied lightweight adapters for image-to-video transfer in action recognition. However, to the best of our knowledge, the potential of PEFT for the complex, dual-objective (spatial and temporal) task of STVG remains unexplored. We go beyond simply applying existing PEFT methods by designing a suite of specialized adapters and a novel temporal decoder to effectively bridge the gap between static image understanding and dynamic video grounding.

%% file: sec/2_method.tex
\section{Method}

\subsection{The Grounding DINO Framework}
Building upon large-scale pre-trained vision-language models, Grounding DINO \cite{liu2024grounding} provides a robust foundation for open-vocabulary detection. It accurately localizes objects described by arbitrary text queries through tight modality fusion, integrating a visual backbone (Swin Transformer \cite{liu2021swin}) and a text encoder. Its language-guided query selection mechanism dynamically refines multimodal encoder outputs into a language-specific ``memory". This refined memory then guides a cross-modality decoder to progressively localize objects, culminating in precise bounding box and text matching for open-vocabulary detection. Grounding DINO's strong generalization and robust performance across diverse scenarios makes it an ideal candidate for spatio-temporal video grounding (STVG) adaptation.

\par Specifically, Grounding DINO's ability to perform open-vocabulary detection, unconstrained by a fixed set of object categories, aligns perfectly with the STVG task. In STVG, the goal is to ground an input textual query, which can be any descriptive phrase, within a video's temporal and spatial dimensions. This intrinsic flexibility of Grounding DINO's open-vocabulary nature makes it exceptionally well-suited for tasks where the target objects or events are described by natural language, rather than limited to predefined classes.

\subsection{Architecture Overview}

\begin{figure*}[htbp] 
  \centering
  \vspace{-1mm}
  \includegraphics[width=0.97\textwidth]{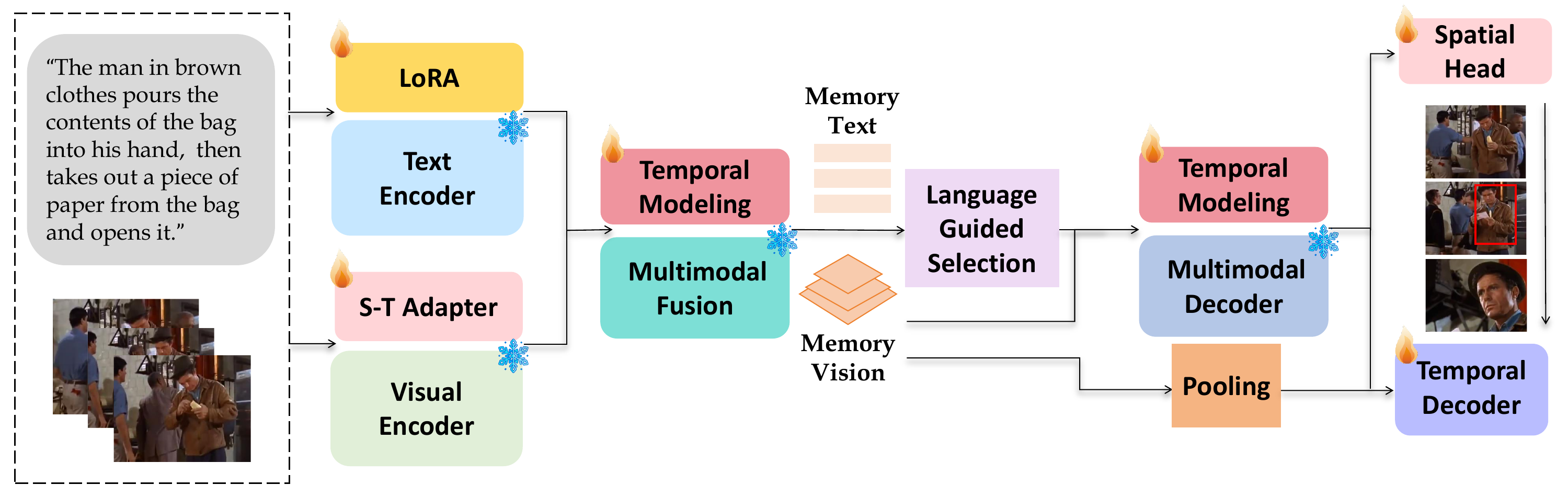} 
  \caption{The proposed Spatial-Temporal Grounding DINO (ST-GD) adapts Grounding DINO for STVG via PEFT, efficiently integrating pre-trained and trainable modules. It processes video and text through fine-tuned visual and language branches, fusing features with temporal modeling. Language-guided selection identifies relevant video features, which are decoded to refine temporal dynamics, generate spatial bounding boxes, and predict temporal boundaries via a dedicated temporal decoder.
 }
 \vspace{-1mm}
  \label{fig:architecture}
\end{figure*}

As illustrated in Fig.\ref{fig:architecture}, our framework, ST-GD, adapts the powerful pre-trained Grounding DINO for spatial-temporal video grounding (STVG). This adaptation addresses the inherent limitations of Grounding DINO, which was initially designed for static images, specifically overcoming its lack of temporal modeling and its high vulnerability to severe overfitting if fully fine-tuned on limited video data. To overcome these challenges, ST-GD incorporates several key innovations:
\begin{itemize}
    \item \textbf{Parameter-Efficient Encoder Adaptation:} By injecting S-T Adapters into the visual encoder and applying LoRA to the text encoder, we efficiently extract visual and text features with minimal parameters.
    \item \textbf{Enhanced Temporal Modeling:} We augment Grounding DINO's core multimodal fusion and decoder components with novel modules for temporal modeling, enabling them to generate intermediate representations rich in spatio-temporal-textual context.
    \item \textbf{Query-Guided Feature Refinement:} Our core innovation leverages Grounding DINO's language-guided query selection mechanism to filter and refine these intermediate features, followed by pooling strategies to aggregate them into a compact representation that serves as the basis for temporal localization.
    \item \textbf{Dedicated Temporal Decoder:} A novel Temporal Decoder refines these processed features to accurately predict temporal boundaries, while inherited spatial heads handle spatial predictions.
\end{itemize}

\subsection{Adaptation for Visual and Language Encoder}
Our S-T Adapter efficiently injects spatio-temporal awareness into the visual encoder, particularly within the context of architectures like Swin Transformer \cite{liu2021swin}. The adapter's design is motivated by the need to capture appearance and motion cues efficiently. It employs spatial and temporal convolutions to effectively modeling spatial features and temporal dynamics.

\par Specifically, the adapter operates on the visual encoder's feature maps. Let $\boldsymbol{Z} \in \mathbb{R}^{T \times H \times W \times C}$ denote the input feature map from a visual encoder stage, where $T, H, W, C$ are time, height, width, and channel dimensions, respectively. The adapter first expands the feature dimension using an up-projection matrix $\boldsymbol{W}_{up}$. Subsequently, it processes these expanded features through parallel spatial and temporal branches: a 2D convolution captures frame-level spatial details, while a 1D convolution models temporal evolution across frames. These parallelly processed features are fused, creating a unified spatio-temporal representation, which is then projected back to the original channel dimension using a down-projection matrix $\boldsymbol{W}_{down}$.

\par Strategically placed at the end of the visual encoder stages, the S-T Adapter uses a shortcut connection \cite{he2016deep,pan2022st} to process global feature maps. Crucially, its weights are zero-initialized to act as an identity mapping initially. In small-data regimes, this prevents immediate disruption of established visual priors, allowing a gradual injection of temporal awareness without catastrophic forgetting on limited data. Operationally, the input's channel dimension is first expanded via a linear layer. It then splits into two parallel branches: a 2D convolution captures intra-frame spatial details, while a 1D convolution (following spatial pooling) models inter-frame temporal dynamics. These branch outputs are fused element-wise, projected back to the original dimension, and integrated via the residual connection. This efficiently injects spatio-temporal awareness while strictly safeguarding pre-trained knowledge.

\par Similarly, while Grounding DINO’s text encoder is powerful, it requires adaptation for the specific vocabulary and sentence structures describing actions in videos. We address this using LoRA (Low-Rank Adaptation) \cite{hu2022lora}. Instead of fully fine-tuning the text encoder—which easily overfits to the narrow vocabulary of small STVG datasets—LoRA provides a data-efficient adaptation that maintains broad semantic comprehension.

\subsection{Temporal Modeling for Multi-modal Modules}
Grounding DINO effectively grounds images to text queries, excelling at static object localization. However, directly applying it to STVG is challenging due to its inherent lack of temporal modeling, where frame-independent attention mechanisms fail to capture video dynamics. Furthermore, explicitly training complex temporal attention modules from scratch, as in prior works \cite{wasim2024videogrounding, pan2022st}, is notoriously data-hungry and highly prone to severe overfitting on small video datasets.

\begin{figure}[htb]
  \centering
  \begin{subfigure}[b]{0.20\textwidth}
    \centering
    \includegraphics[width=0.9\linewidth]{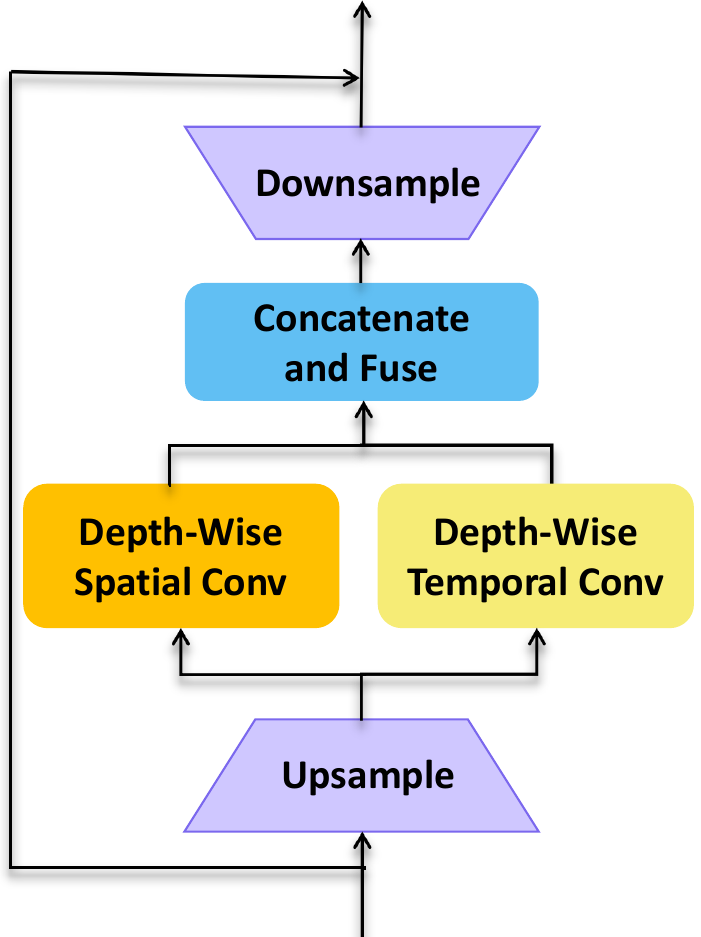}
    \caption{S-T Adapter}
    \label{fig:subfig_a}
  \end{subfigure}%
  \hspace{0.4em}
  \begin{subfigure}[b]{0.16\textwidth}
    \centering
    \includegraphics[width=0.80\linewidth]{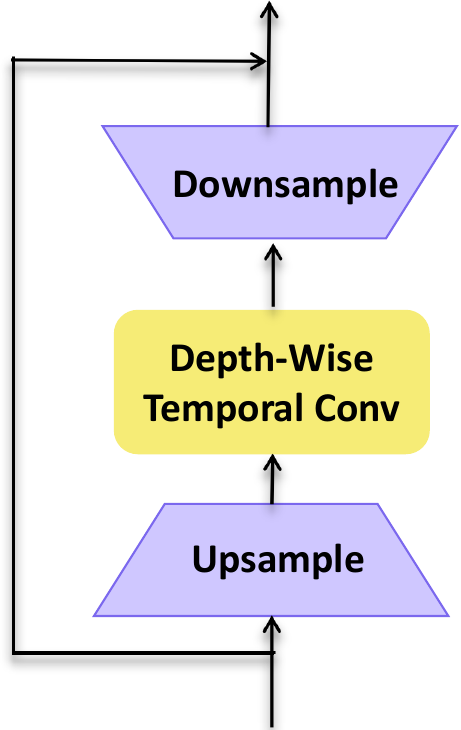}
    \caption{Temporal Adapter}
    \label{fig:subfig_b}
  \end{subfigure}%

  \vspace{0.01em}
  \begin{subfigure}[b]{0.40\textwidth}
    \centering
    \includegraphics[width=0.90\linewidth]{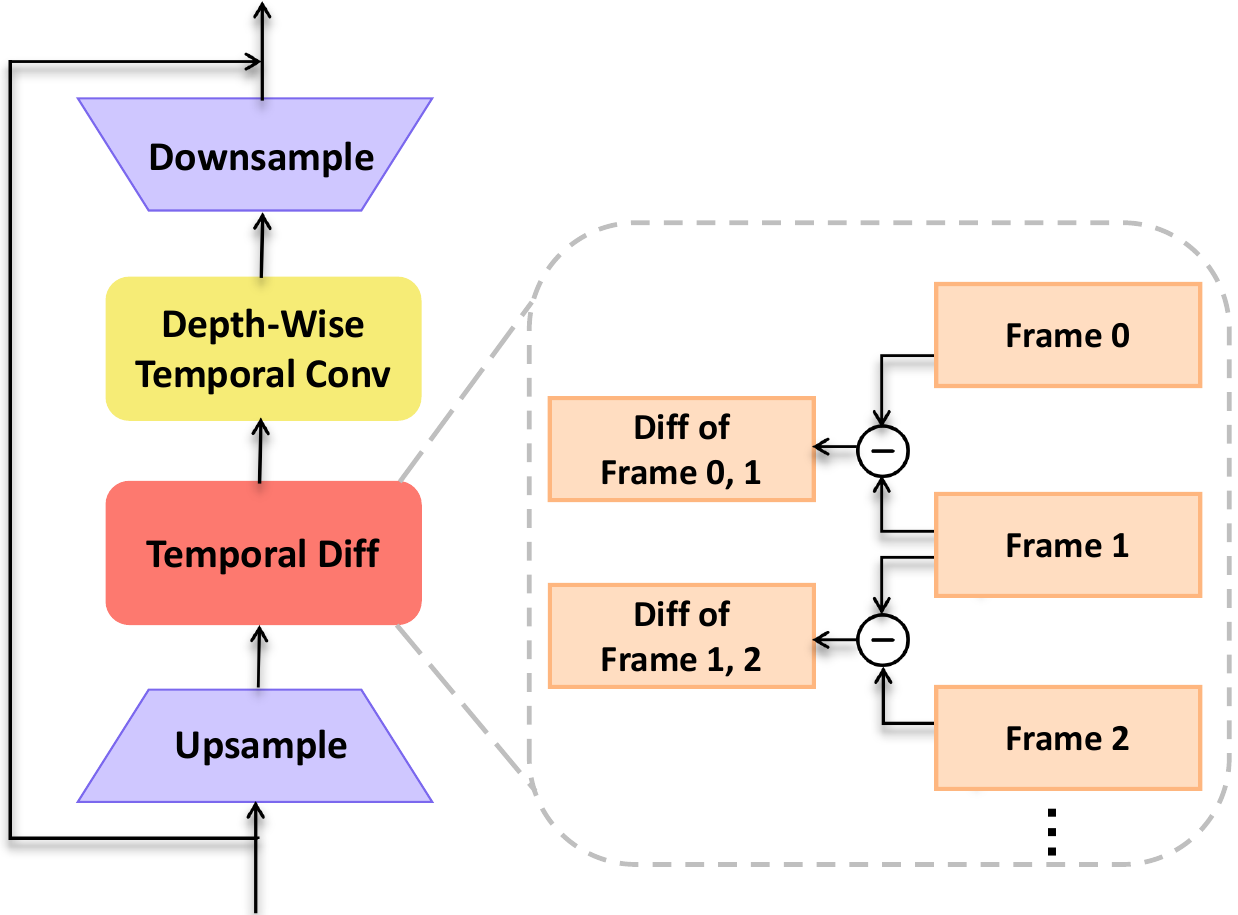}
    \caption{Temporal Difference Adapter}
    \label{fig:subfig_c}
  \end{subfigure}

  \caption{To better transfer Grounding DINO to STVG task, three different adapters following PEFT style are proposed to model temporal relationships between time steps.}
  \label{fig:main_figure}
  \vspace{-0.5em}
\end{figure}

To efficiently address these limitations, our ST-GD framework integrates lightweight Temporal Adapters into the multimodal encoder and decoder. This design aims to capture inherent temporal dependencies and enhance query refinement. Similar to the S-T Adapter, the Temporal Adapter, as illustrated in Fig. \ref{fig:subfig_b}, processes an input feature $\boldsymbol{Z} \in \mathbb{R}^{T \times D}$ (where $T$ is temporal dimension and $D$ is feature dimension) by applying a 1D convolution along the temporal axis. The adapter can be expressed as:
\begin{equation}
\boldsymbol{Z}_{out} = \boldsymbol{Z} + \text{Conv1d}(\boldsymbol{Z}\boldsymbol{W}_{up})\boldsymbol{W}_{down}.
\label{eq:temporal_adapter}
\end{equation}
To explicitly model fine-grained temporal dynamics and precisely predict temporal boundaries, we introduce the Temporal-Diff Adapter. Its core motivation stems from the observation that an event's critical temporal start and end points often coincide with significant shifts or transitions in video content, manifesting as divergence in semantic and visual features between adjacent frames. Our Temporal-Diff Adapter, as illustrated in Fig. \ref{fig:subfig_c}, is designed to capture these frame-to-frame dissimilarities directly. It calculates the difference between feature representations of adjacent frames, generating ``temporal difference" representation that highlights regions of high temporal variability. This explicit modeling of feature divergence is crucial for accurately identifying precise start and end frames. To capture these frame-to-frame dissimilarities, we define a temporal difference operator, $\Delta_t(\cdot)$, where $\Delta_t(\boldsymbol{Q})_t = \boldsymbol{Q}_{t+1} - \boldsymbol{Q}_t$ for a sequence $\boldsymbol{Q}$. We apply zero-padding to the final timestep to maintain the sequence length. The adapter can then be expressed as:
\begin{equation}
\boldsymbol{Q}_{\text{out}} = \boldsymbol{Q} + \text{Conv1D}\left(\Delta_t(\boldsymbol{Q}) \boldsymbol{W}_{\text{up}}\right) \boldsymbol{W}_{\text{down}}.
\label{eq:temporal_diff_adapter_revised}
\end{equation}

\begin{itemize}
    \item \textbf{Encoder-side Temporal Adapter:} To infuse temporal awareness into Grounding DINO’s multimodal fusion process, we apply Temporal Adapters to the visual features prior to their bi-attention fusion with text features within the encoder. This ensures the encoder’s output, comprising `Memory Text' and `Memory Vision', is inherently rich in spatio-temporal context. Subsequently, these refined memories are further distilled via Grounding DINO’s language-guided selection, which provides precise guidance for spatial localization.

    \item \textbf{Decoder-side Temporal Adapter:} Building upon temporally-enhanced features, Temporal Adapters are integrated into the multimodal decoder, specifically before key cross-modal interaction modules. This refines temporal modeling during query refinement and models relationships between queries of different frames.

    \item \textbf{Enhanced Temporal Modeling via Temporal Difference Adapter:} The Temporal Difference Adapter is strategically integrated at critical junctures to model frame-to-frame feature changes. It is applied before and after the main multimodal fusion modules to provide stronger temporal signals and capture dynamic changes in combined representations. Furthermore, a Temporal Difference Adapter is employed after the decoder to refine spatial features for temporal grounding.
\end{itemize}

\subsection{Query-Guided Feature Refinement}
Our core innovation for precise temporal localization lies in effectively leveraging and extending Grounding DINO’s successful multi-modal spatial grounding paradigm. We repurpose its powerful language-guided query selection mechanism to derive robust, text-aligned frame representations for temporal localization.

Specifically, from the final multimodal decoder layer, we extract the refined visual queries $\boldsymbol{V} \in \mathbb{R}^{T \times N_q \times d}$ and the fused text query feature $\boldsymbol{c} \in \mathbb{R}^{d}$ (e.g. the \texttt{[CLS]} token).

First, we compute the relevance score $s_{t,j}$ for the $j$-th visual query in frame $t$, denoted as $\boldsymbol{v}_{t,j} \in \mathbb{R}^d$, by its cosine similarity to the text query:
\begin{equation}
s_{t,j} = \frac{\boldsymbol{v}_{t,j} \cdot \boldsymbol{c}}{\|\boldsymbol{v}_{t,j}\| \|\boldsymbol{c}\|}.
\label{eq:similarity_score}
\end{equation}

Next, for each frame $t$, we identify the set of indices $\mathcal{K}_t$ corresponding to the top-$K$ most relevant queries.

Finally, these top-$K$ queries are aggregated into a single frame-level feature $\boldsymbol{f}_t^{\text{agg}}$ via a weighted sum. The weights are derived from the normalized relevance scores of the selected queries:
\begin{equation}
\alpha_{t,k} = \frac{\exp(s_{t,k})}{\sum_{j \in \mathcal{K}_t} \exp(s_{t,j})}, \quad \forall k \in \mathcal{K}_t,
\label{eq:softmax_weights}
\end{equation}
\begin{equation}
\boldsymbol{f}_t^{\text{agg}} = \sum_{k \in \mathcal{K}_t} \alpha_{t,k} \cdot \boldsymbol{v}_{t,k}.
\label{eq:aggregation}
\end{equation}
This vector $\boldsymbol{f}_t^{\text{agg}}$ serves as a robust, text-aligned local representation for each frame. It is then fused with global features, extracted from the encoder's output, to form a comprehensive frame representation for the subsequent temporal decoder.

\subsection{Dedicated Temporal Decoder}
With these robustly refined and contextually rich frame-level features and their associated confidence scores, we design a novel Temporal Decoder to predict temporal boundaries accurately. While preceding convolutional operations (like those in our adapters) effectively capture local temporal dependencies, predicting global event boundaries requires understanding long-range temporal relationships, which attention mechanisms are inherently well-suited for.

\par The core of our Temporal Decoder is an enhanced temporal attention mechanism. The comprehensive frame representations (combining local query and global features) along with relative positional encodings are fed into this module. Unlike traditional temporal convolutions with limited receptive fields, the temporal attention mechanism can effectively model relationships between distant frames across the entire video. This is crucial because event boundaries are defined by changes that might occur non-locally in time. By integrating these elements, our Temporal Decoder refines temporal boundary predictions, enabling the precise localization of the start and end frames of the described event.

\begin{table*}[htb]

 \caption{Comparison with other methods on the limited-scale HCSTVG-v1 test set (\%). As a severely data-scarce benchmark, our data-efficient PEFT approach naturally prevents overfitting and achieves highly competitive results.}

  \centering
    \begin{tabular}{lcccccc}
    \toprule
    Methods & Training Method & m\_tIoU & m\_vIoU & vIoU@0.3 & vIoU@0.5  & TP\\
    \midrule
    STVGBert [ICCV2021]\cite{su2021stvgbert} & Trained from Scratch & -     & 20.4  & 29.4  & 11.3 & - \\
    TubeDETR [CVPR22]\cite{yang2022tubedetr} & Trained from Scratch & 43.7  & 32.4  & 49.8  & 23.5 & 185M \\
    STCAT [NeurIPS22]\cite{jin2022embracing} & Trained from Scratch & 49.4  & 35.1  & 57.7  & 30.1 & 207M \\
    CSDVL [CVPR23]\cite{lin2023collaborative} & Trained from Scratch & -     & 36.9  & 62.2  & 34.8 & - \\
    CG-STVG[CVPR24]\cite{gu2024context} & Trained from Scratch & \underline{52.8} & 38.4  & 61.5  & 36.3 & 203M \\
    VideoGrounding-DINO[CVPR24]\cite{wasim2024videogrounding}   & Partially Fine-tuned & -  &  38.3  &  62.4  &   36.1 & - \\
    \midrule
    E3MP[ECCV24]\cite{bao2024e3m}  & Zero Shot   & - & 19.1 & 29.4 & 10.6 & -\\
    CVTP[EMNLP24]\cite{zhao2024video} & Weakly Supervised & - & 16.4 & 18.7 & 8.3 \\
    STPro[CVPR25]\cite{garg2025stpro}  & Weakly Supervised & - & 17.6 & 27.0 & 12.9 & -\\
    RealVG[ACM MM25]\cite{wei2025realvg} & Zero Shot & - & 29.5 & 40.0 & 25.8 & - \\
\midrule
\rowcolor[RGB]{222,236,215} 
    ST-GD(Swin-T)  & PEFT & 52.0  &  \underline{39.2}  &  \underline{62.5}  &   \underline{37.5}  & \textbf{9.4M}\\
\rowcolor[RGB]{222,236,215} 
    ST-GD(Swin-B)  & PEFT & \textbf{53.1}  &  \textbf{39.9}  &  \textbf{65.1}  &   \textbf{38.4}  & \underline{10.0M}\\
    \bottomrule
    \end{tabular}%
  \label{tab:v1}%
\end{table*}%

\subsection{Optimization}
Given a video and its text queries, our model predicts:
\begin{enumerate}
    \item Start timestamps $H_s^k = \{h_s^i\}_{i=1}^{N}$ and end timestamps $H_e^k = \{h_e^i\}_{i=1}^{N}$ of the video clip related to text, $N$ is the number of total frames.
    \item Bounding box $B^k = \{b^i\}_{i=1}^{N}$ of the referred object.
    \item Temporal and query confidence scores $s_t^k$, $s_q^k$.
\end{enumerate}

In training, given ground truth start timestamps $H_s^*$, the end timestamps $H_e^*$, and the bounding box sequence $B^*$. For temporal grounding, the KL divergence and binary cross-entropy are used as the loss function, and the losses of start and end times are computed as follows:
\begin{align} 
    L_t^k =& \lambda_s L_{KL}(H_s^*, H_s^k) + \lambda_e L_{KL}(H_e^*,\nonumber H_e^k)\\ &+ \lambda_{t} L_{BCE}((H_s^*, H_e^*), s_t^k).
\end{align}
For the spatial grounding objective, our model inherits the loss formulation from DETR-like architectures \cite{zhu2020deformable}. The loss is computed only for the frames within the ground-truth temporal segment $[t_s, t_e]$. The spatial loss, $\mathcal{L}_{\text{s}}$, is a linear combination of the smooth $L_1$ loss and the Generalized Intersection over Union (GIoU) loss:
\begin{equation}
\mathcal{L}_{\text{s}} = \lambda_{box} \lambda_{\text{box}}(\hat{\boldsymbol{B}}, \boldsymbol{B}) + \lambda_{\text{giou}} \mathcal{L}_{\text{giou}}(\hat{\boldsymbol{B}}, \boldsymbol{B}),
\label{eq:spatial_loss}
\end{equation}

The total training loss for training is:
\begin{equation}
    L = \sum_{k=1}^{K} (L_t^k + L_s^k).
\end{equation}

\begin{table*}[ht]
\centering
\caption{Performance comparisons on the VidSTG test set (\%). Despite VidSTG featuring a significantly larger scale and highly complex relational queries, our lightweight method maintains robust generalization against fully-trained models.}

\label{tab:vidstg_results_booktabs}
\begin{tabular}{l cccc cccc}
\toprule
\multirow{2}{*}{Methods} & \multicolumn{4}{c}{Declarative Sentences} & \multicolumn{4}{c}{Interrogative Sentences} \\
\cmidrule(lr){2-5} \cmidrule(lr){6-9}
& m\_tIoU & m\_vIoU & vIoU@0.3 & vIoU@0.5 & m\_tIoU & m\_vIoU & vIoU@0.3 & vIoU@0.5 \\
\midrule
TubeDETR [CVPR22]\cite{yang2022tubedetr}  & 48.1 & 30.4 & 42.5 & 28.2 & 46.9 & 25.7 & 35.7 & 23.2 \\
STCAT [NeurIPS22]\cite{jin2022embracing}  & 50.8 & 33.1 & 46.2 & 32.6 & 49.7 & 28.2 & 39.2 & 26.6 \\
CSDVL [CVPR23]\cite{lin2023collaborative}  & - & 33.7 & 47.2 & 32.8 & - & 28.5 & 39.9 & 26.2 \\
CG-STVG[CVPR24]\cite{gu2024context}   & 51.4 & 34.0 & 47.7 & 33.1 & 49.9 & 29.0 & 40.5 & 27.5 \\
VideoGrounding-DINO\cite{wasim2024videogrounding}  & \textbf{51.9} & \underline{34.6} & \underline{48.1} & \textbf{33.9} & \underline{50.8} & 29.8 & \underline{41.0} & \underline{27.6}\\
    \midrule
    E3MP[ECCV24]\cite{bao2024e3m}  & - & 16.2 & 20.4 & 11.9 & - & 10.6 & 12.2 & 5.4 \\
    CVTP[EMNLP24]\cite{zhao2024video} & - & 17.9 & 22.3 & 14.9 & - & 11.2 & 12.4 & 7.2 \\
    STPro[CVPR25]\cite{garg2025stpro} & -  & 15.5 & 19.4 & 12.6 & - & 12.5 & 14.9 & 9.3\\
    RealVG[ACM MM25]\cite{wei2025realvg}  & 37.4 & 29.0 & 34.9 & 25.5 & 36.7 & \underline{30.0} & 34.7 & 25.7 \\
\midrule
\rowcolor[RGB]{222,236,215} 
ST-GD(Swin-T) & 50.2 & 33.6 & 45.7 & 32.8 & 49.6 & 28.7  & 39.4 & 27.3 \\
\rowcolor[RGB]{222,236,215} 
ST-GD(Swin-B) & \underline{51.5} & \textbf{34.8} & \textbf{48.5} & \underline{33.4} & \textbf{51.2} & \textbf{30.3}  & \textbf{41.2} & \textbf{28.8} \\
\midrule
\bottomrule
\end{tabular}
\label{tab:vidstg}%
\vspace{-0.5em}
\end{table*}

%% file: sec/3_experiment.tex
\section{Experiments}

\begin{table}[htbp]
  \caption{Comparison on HCSTVG-v2 test set (\%).}
  \vspace{-0.5em}
  \centering
  \resizebox{1.0\linewidth}{!}{
    \begin{tabular}{lcccc}
    \toprule
    Methods  & m\_tIoU & m\_vIoU & vIoU@0.3 & vIoU@0.5 \\
    \midrule

    TubeDETR[CVPR22]\cite{yang2022tubedetr}    & 53.9   & 36.4  & 58.8  & 30.6 \\
    CSDVL[CVPR23]\cite{lin2023collaborative}   & 58.1  & 38.7  & 65.5  & 33.8 \\
    CGSTVG[CVPR24]\cite{gu2024context}  & 59.2 & 39.5  & 61.5 &36.3   \\

    \midrule
    STPro[CVPR25]\cite{garg2025stpro}   & - & 20.0 & 31.7 & 14.6 \\
    RealVG[ACM MM25]\cite{wei2025realvg}  & - & 33.1 & 42.3 & 27.0  \\
    \midrule
    \rowcolor[RGB]{222,236,215} 
    ST-GD(Swin-T)   & 58.1 &  38.4  &  63.4  &   34.7 \\
    \rowcolor[RGB]{222,236,215} 
    ST-GD(Swin-B)    & 59.2 &  \textbf{40.1}  &   \underline{66.5}  &   \textbf{37.3}  \\
    \bottomrule
    \end{tabular} }
  \label{tab:v2}%
  \vspace{-0.4em}
\end{table}%

\subsection{Implementation}
Our approach is built upon the Grounding DINO \cite{author_title_year} model, utilizing its Swin Transformer-based \cite{liu2021swin} image backbone and BERT-based \cite{DBLP:journals/corr/abs-1810-04805} text backbone. Grounding DINO offers two official checkpoints, Swin-T and Swin-B, representing different scales. Our experiments comprehensively evaluate both to understand their strengths and limitations in spatial-temporal video grounding.

\subsection{Datasets}
The HC-STVG\cite{tang2021human} v1 and v2 datasets are crucial benchmarks for evaluating spatial-temporal video grounding models, specifically focusing on the challenging task of localizing described individuals within complex, multi-person scenes. Additionally, we use the VidSTG\cite{zhang2020does} dataset to evaluate performance on queries involving intricate object relationships and interrogative sentences.

\subsection{Metrics}
Following Tubedetr \cite{yang2022tubedetr}, STCAT \cite{jin2022embracing}, and CGSTVG \cite{gu2024context}, we employ mean temporal Intersection-over-Union (m tIoU), mean video Intersection-over-Union (m vIoU), and vIoU@R as our primary evaluation metrics. We compare trainable parameters (TP) against existing approaches to highlight our method's efficiency. ST-GD’s TP, including its adapters, temporal decoder, spatial head, and temporal head, demonstrates significant complexity reduction.

\subsection{Result Comparison}

Results on these benchmarks highlight ST-GD’s effectiveness and  parameter efficiency. Unlike methods like Video-GroundingDINO\cite{wasim2024videogrounding} (partial fine-tuning) or CGSTVG\cite{gu2024context} (full fine-tuning), ST-GD exclusively uses PEFT by keeping the Grounding DINO backbone frozen and introducing only lightweight adapters for maximum efficiency. On the smaller HC-STVG (v1) dataset, ST-GD demonstrates SOTA performance, underscoring PEFT’s efficiency under limited resources. On the other end of the spectrum are zero-shot and weakly supervised methods such as E3MP\cite{bao2024e3m} and RealVG\cite{wei2025realvg}. Although these methods leverage the powerful general knowledge of massive vision-language models (e.g., Qwen-VL-7B\cite{wang2024qwen2}), they exhibit a significant performance gap on this specialized task. They struggle to adapt their broad understanding to the fine-grained spatio-temporal localization required by STVG.
For the larger, more complex HCSTVG-v2 benchmark, ST-GD exhibits more balanced and robust results, highly competitive despite significantly lower parameter count. This balanced performance, a key strength of our PEFT approach, offers superior generalization and a more consistent profile than models with skewed scores. This is particularly advantageous for limited-data scenarios with high annotation costs. 

We further evaluate ST-GD on the biggest VidSTG benchmark, which features a wider variety of objects and complex relational queries, including interrogative sentences. As shown in Table \ref{tab:vidstg}, our PEFT-based approach remains highly competitive against state-of-the-art methods that employ full fine-tuning or partial fine-tuning strategies.

This robust performance on a complex, multi-category dataset, achieved with significantly fewer trainable parameters, highlights the primary advantage of our framework: ST-GD effectively transfers the strong generalization capabilities of the Grounding DINO foundation model to a demanding video-language task, demonstrating superior efficiency without sacrificing performance.

 \begin{figure*}[t] 
  \centering
  \vspace{-1em}
  \includegraphics[width=0.9\linewidth]{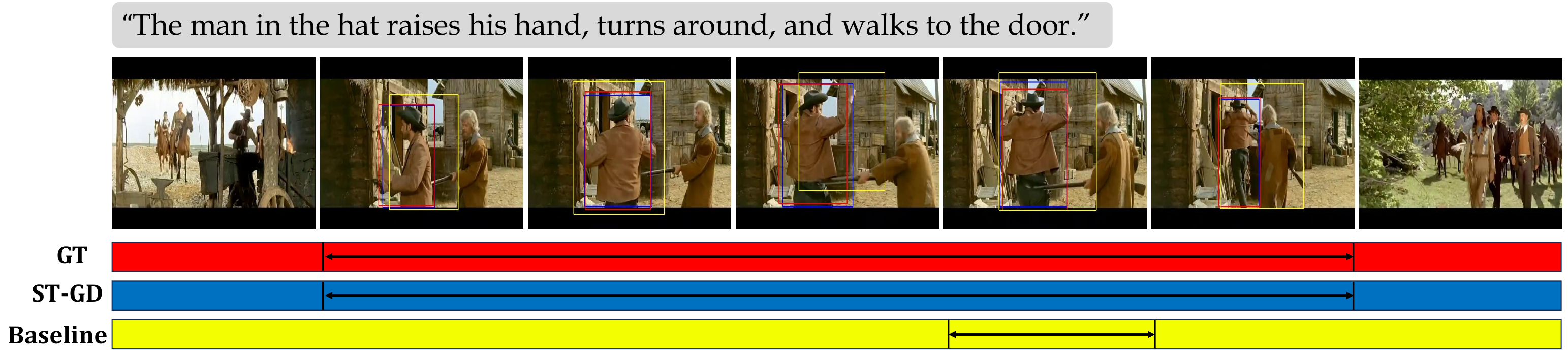} 
  \caption{Visualization results on the HCSTVG-v1 test set show that ST-GD (blue) outperforms the partially fine-tuned Grounding DINO baseline (yellow) by better understanding spatial-temporal interactions.
 }
  \label{fig:temporal_vis}
\end{figure*}

\begin{figure}[t] 
  \centering
  \vspace{-1em}
  \includegraphics[width=0.99\linewidth]{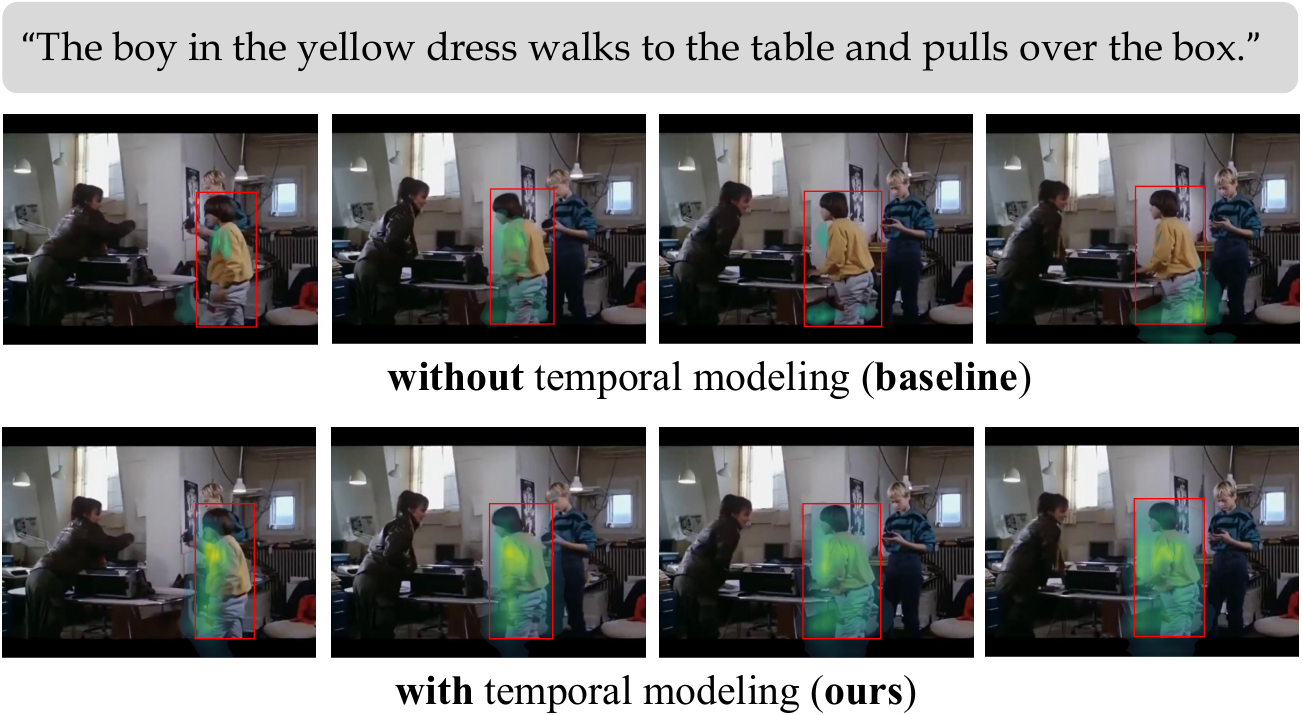}
  \caption{Attention maps on the HCSTVG-v1 test set highlight the importance of temporal modeling for capturing target dynamics and relationships.
 }
  \label{fig:spatial_vis}

\end{figure}

\subsection{Ablation Experiments}
We conduct extensive ablation studies on the HC-STVG v1 dataset using Swin-T Grounding DINO's backbone to analyze the contribution of each proposed component within ST-GD. The results are summarized in Table \ref{tab:ablation_combined}, \ref{tab:ablation_study}.

\noindent\textbf{Effectiveness of Proposed Components.}
Table \ref{tab:ablation_combined} dissects the contributions of our tailored adapters and the Temporal Decoder. As shown in Table \ref{tab:ablation_combined}(a), replacing our specialized modules with naive 1D temporal convolutions causes a significant performance decline. The S-T Adapter integrates comprehensive spatio-temporal features, while the Temporal-Diff Adapter explicitly captures inter-frame dynamics; their complementary roles are essential for robust temporal modeling. Furthermore, Table \ref{tab:ablation_combined}(b) validates the Temporal Decoder. The Fusion module effectively integrates global encoder outputs with local decoder features, while the Aggregation module leverages language-guided selection to weight queries based on text relevance. Removing either mechanism degrades boundary prediction accuracy. Together, these carefully designed, lightweight components enable precise temporal grounding while maintaining strict parameter and data efficiency.

\begin{table}[htbp]
    \centering
    \caption{Ablations on the proposed components. ST-GD represents our full data-efficient model.}
    \vspace{-0.2em}
    \resizebox{1.0\linewidth}{!}{
    \begin{tabular}{lcccc}
        \toprule
        Ablation Setting & m\_tIoU & m\_vIoU & vIoU@0.3 & vIoU@0.5  \\ 
        \midrule
        \textbf{ST-GD (Full Model)} & \textbf{52.0}  &  \textbf{39.2}  &  \textbf{62.5}  &   \textbf{37.5}   \\
        \midrule
        \multicolumn{5}{l}{\textit{(a) Ablation on Adapters}} \\
        S-T  $\rightarrow$ T Adapter & 50.5  &  36.7  &  57.6  &   34.1  \\
        T-D  $\rightarrow$ T Adapter & 51.3  &  38.0  &  59.7  &   34.5  \\
        \midrule
        \multicolumn{5}{l}{\textit{(b) Ablation on Temporal Decoder}} \\
        Without Fusion & 51.2  &  38.2  &  60.4  &   36.5  \\
        Without Aggregation & 51.4  &  38.5 &   60.9 &   36.7  \\
        \bottomrule
    \end{tabular}}
    \label{tab:ablation_combined}

\end{table}

\begin{table}[htbp]
    \centering
      \caption{Ablation study on the contribution of components. }
      \vspace{-0.2em}
     \resizebox{1.0\linewidth}{!}{
    \begin{tabular}{lccccc}
        \hline
        Ablation Setting & m\_tIoU & m\_vIoU & vIoU@0.3 & vIoU@0.5  & TP \\ 
        \hline
        Baseline & 31.8  &  23.6  &  31.0  &   14.6  & 35.7M\\
        \hline
        Frozen Grounding DINO \\
        + T Decoder & 45.2 & 30.7 & 47.2 & 22.1 & 3.3M\\
        + T-D Adapter & 46.1 & 32.7 & 51.2 & 26.8 &3.8M\\
        + T Adapter & 49.9 & 36.9 & 60.0 &34.6 &8.3M\\
        
        \hline
        + S-T Adapter + LoRA & 52.0  &  39.2  &  62.5  &   37.5  & 9.4M \\
        \hline
    \end{tabular}}
    \label{tab:ablation_study}
    \vspace{-0.5em}
\end{table}

\noindent\textbf{Contribution of Each Component.}
Table \ref{tab:ablation_study} progressively reveals each component’s impact on performance and trainable parameters. We set partially fine-tuned Grounding DINO with frozen visual and text encoder and adding spatial and temporal heads as a stronger baseline than the minimal baseline used in Figure \ref{fig:compare}, which highlights static image models’ inadequacy for video grounding, emphasizing the need for temporal modeling. Adding a Temporal Decoder to frozen Grounding DINO significantly boosts performance, proving its critical role. Integrating Temporal-Diff Adapters further improves results by capturing event boundaries. Incorporating Temporal Adapters into the Spatial Decoder and Multimodal Fusion enhances performance, infusing temporal awareness into spatial localization and multi-modal fusion. Finally, the S-T Adapter and LoRA for PEFT consolidate the full model’s superior performance, providing comprehensive spatio-temporal learning with efficiency. These ablations collectively validate each component’s necessity and effectiveness for ST-GD’s enhanced spatio-temporal localization.

\subsection{Qualitative Results}
We trained ST-GD and a baseline Grounding DINO (with partially fine-tuned) on HCSTVG-v1, evaluating performance on its test set. Fig. \ref{fig:temporal_vis} visualizes spatio-temporal attention map with targets bounding boxes: the baseline showed inconsistent spatial and limited temporal grounding, often misinterpreting periods. Fig. \ref{fig:spatial_vis} visualizes attention maps; the baseline’s top row is inconsistent, while our method’s (bottom row) effectively captures target dynamics for robust spatio-temporal grounding.

%% file: sec/4_conclusion.tex
\section{Conclusion}

We presented ST-GD, a framework that successfully adapts Grounding DINO for STVG using a novel PEFT strategy. By integrating specialized adapters and a dedicated temporal decoder, our method efficiently learns temporal dynamics. On the HC-STVG and VidSTG benchmarks, ST-GD establishes a superior performance-efficiency trade-off: it achieves competitive results against full or partial fine-tuning with a fraction of the parameters, while significantly outperforming zero-shot and weakly supervised methods. However, we acknowledge the frozen-backbone assumption may limit performance on complex datasets by potentially hindering the learning of novel video-specific patterns. Our work offers a promising direction for efficiently leveraging foundation models for dynamic video tasks. We hope our findings encourage further research into data-efficient adaptation via PEFT, offering a practical pathway to scale video-language understanding in domains where annotated data is inherently scarce.

\clearpage
\section{Acknowledgment}
This work was supported by the National Natural Science Foundation of China under Grant No. 62403429 and Zhejiang Province Natural Science Foundation of China under Grant No. LQN25F030008.

%% file: main.bib
@String(ICLR = {Int. Conf. Learn. Represent.})

@String(ICLR  = {ICLR})

@inproceedings{wasim2024videogrounding,
  title={Videogrounding-dino: Towards open-vocabulary spatio-temporal video grounding},
  author={Wasim, Syed Talal and Naseer, Muzammal and Khan, Salman and Yang, Ming-Hsuan and Khan, Fahad Shahbaz},
  booktitle={Proceedings of the IEEE/CVF Conference on Computer Vision and Pattern Recognition},
  pages={18909--18918},
  year={2024}
}

@article{zhu2020deformable,
  title={Deformable detr: Deformable transformers for end-to-end object detection},
  author={Zhu, Xizhou and Su, Weijie and Lu, Lewei and Li, Bin and Wang, Xiaogang and Dai, Jifeng},
  journal={arXiv preprint arXiv:2010.04159},
  year={2020}
}

@article{hu2022lora,
  title={Lora: Low-rank adaptation of large language models.},
  author={Hu, Edward J and Shen, Yelong and Wallis, Phillip and Allen-Zhu, Zeyuan and Li, Yuanzhi and Wang, Shean and Wang, Lu and Chen, Weizhu and others},
  journal={ICLR},
  volume={1},
  number={2},
  pages={3},
  year={2022}
}

@inproceedings{zhang2020does,
  title={Where does it exist: Spatio-temporal video grounding for multi-form sentences},
  author={Zhang, Zhu and Zhao, Zhou and Zhao, Yang and Wang, Qi and Liu, Huasheng and Gao, Lianli},
  booktitle={Proceedings of the IEEE/CVF Conference on Computer Vision and Pattern Recognition},
  pages={10668--10677},
  year={2020}
}

@article{pan2022st,
  title={St-adapter: Parameter-efficient image-to-video transfer learning},
  author={Pan, Junting and Lin, Ziyi and Zhu, Xiatian and Shao, Jing and Li, Hongsheng},
  journal={Advances in Neural Information Processing Systems},
  volume={35},
  pages={26462--26477},
  year={2022}
}

@inproceedings{liu2024grounding,
  title={Grounding dino: Marrying dino with grounded pre-training for open-set object detection},
  author={Liu, Shilong and Zeng, Zhaoyang and Ren, Tianhe and Li, Feng and Zhang, Hao and Yang, Jie and Jiang, Qing and Li, Chunyuan and Yang, Jianwei and Su, Hang and others},
  booktitle={European Conference on Computer Vision},
  pages={38--55},
  year={2024},
  organization={Springer}
}

@inproceedings{su2021stvgbert,
  title={Stvgbert: A visual-linguistic transformer based framework for spatio-temporal video grounding},
  author={Su, Rui and Yu, Qian and Xu, Dong},
  booktitle={Proceedings of the IEEE/CVF International Conference on Computer Vision},
  pages={1533--1542},
  year={2021}
}

@inproceedings{yang2022tubedetr,
  title={Tubedetr: Spatio-temporal video grounding with transformers},
  author={Yang, Antoine and Miech, Antoine and Sivic, Josef and Laptev, Ivan and Schmid, Cordelia},
  booktitle={Proceedings of the IEEE/CVF Conference on Computer Vision and Pattern Recognition},
  pages={16442--16453},
  year={2022}
}

@article{tang2021human,
  title={Human-centric spatio-temporal video grounding with visual transformers},
  author={Tang, Zongheng and Liao, Yue and Liu, Si and Li, Guanbin and Jin, Xiaojie and Jiang, Hongxu and Yu, Qian and Xu, Dong},
  journal={IEEE Transactions on Circuits and Systems for Video Technology},
  volume={32},
  number={12},
  pages={8238--8249},
  year={2021},
  publisher={IEEE}
}

@inproceedings{cao2022locvtp,
  title={Locvtp: Video-text pre-training for temporal localization},
  author={Cao, Meng and Yang, Tianyu and Weng, Junwu and Zhang, Can and Wang, Jue and Zou, Yuexian},
  booktitle={European Conference on Computer Vision},
  pages={38--56},
  year={2022},
  organization={Springer}
}

@article{jin2022embracing,
  title={Embracing consistency: A one-stage approach for spatio-temporal video grounding},
  author={Jin, Yang and Yuan, Zehuan and Mu, Yadong and others},
  journal={Advances in Neural Information Processing Systems},
  volume={35},
  pages={29192--29204},
  year={2022}
}

@inproceedings{gu2024context,
  title={Context-guided spatio-temporal video grounding},
  author={Gu, Xin and Fan, Heng and Huang, Yan and Luo, Tiejian and Zhang, Libo},
  booktitle={Proceedings of the IEEE/CVF Conference on Computer Vision and Pattern Recognition},
  pages={18330--18339},
  year={2024}
}

@article{tan2021augmented,
  title={Augmented 2d-tan: A two-stage approach for human-centric spatio-temporal video grounding},
  author={Tan, Chaolei and Lin, Zihang and Hu, Jian-Fang and Li, Xiang and Zheng, Wei-Shi},
  journal={arXiv preprint arXiv:2106.10634},
  year={2021}
}

@inproceedings{lin2023collaborative,
  title={Collaborative static and dynamic vision-language streams for spatio-temporal video grounding},
  author={Lin, Zihang and Tan, Chaolei and Hu, Jian-Fang and Jin, Zhi and Ye, Tiancai and Zheng, Wei-Shi},
  booktitle={Proceedings of the IEEE/CVF Conference on Computer Vision and Pattern Recognition},
  pages={23100--23109},
  year={2023}
}

@inproceedings{liu2021swin,
  title={Swin transformer: Hierarchical vision transformer using shifted windows},
  author={Liu, Ze and Lin, Yutong and Cao, Yue and Hu, Han and Wei, Yixuan and Zhang, Zheng and Lin, Stephen and Guo, Baining},
  booktitle={Proceedings of the IEEE/CVF international conference on computer vision},
  pages={10012--10022},
  year={2021}
}

@misc{author_title_year,
  author = {GroundingDINO},
  year = {2024},
  url = {https://github.com/IDEA-Research/GroundingDINO},
}

@article{DBLP:journals/corr/abs-1810-04805,
  author       = {Jacob Devlin and
                  Ming{-}Wei Chang and
                  Kenton Lee and
                  Kristina Toutanova},
  title        = {{BERT:} Pre-training of Deep Bidirectional Transformers for Language
                  Understanding},
  journal      = {CoRR},
  volume       = {abs/1810.04805},
  year         = {2018},
  url          = {http://arxiv.org/abs/1810.04805},
  eprinttype    = {arXiv},
  eprint       = {1810.04805},
  bibsource    = {dblp computer science bibliography, https://dblp.org}
}

@article{wang2022learning,
  title={Learning spatiotemporal and motion features in a unified 2d network for action recognition},
  author={Wang, Mengmeng and Xing, Jiazheng and Su, Jing and Chen, Jun and Liu, Yong},
  journal={IEEE Transactions on Pattern Analysis and Machine Intelligence},
  volume={45},
  number={3},
  pages={3347--3362},
  year={2022},
  publisher={IEEE}
}

@inproceedings{weng2024longvlm,
  title={Longvlm: Efficient long video understanding via large language models},
  author={Weng, Yuetian and Han, Mingfei and He, Haoyu and Chang, Xiaojun and Zhuang, Bohan},
  booktitle={European Conference on Computer Vision},
  pages={453--470},
  year={2024},
  organization={Springer}
}

@inproceedings{he2016deep,
  title={Deep residual learning for image recognition},
  author={He, Kaiming and Zhang, Xiangyu and Ren, Shaoqing and Sun, Jian},
  booktitle={Proceedings of the IEEE conference on computer vision and pattern recognition},
  pages={770--778},
  year={2016}
}

@article{auto1,
  title={A survey of autonomous driving: Common practices and emerging technologies},
  author={Yurtsever, Ekim and Lambert, Jacob and Carballo, Alexander and Takeda, Kazuya},
  journal={IEEE access},
  volume={8},
  pages={58443--58469},
  year={2020},
  publisher={IEEE}
}

@inproceedings{auto2,
  title={Planning-oriented autonomous driving},
  author={Hu, Yihan and Yang, Jiazhi and Chen, Li and Li, Keyu and Sima, Chonghao and Zhu, Xizhou and Chai, Siqi and Du, Senyao and Lin, Tianwei and Wang, Wenhai and others},
  booktitle={Proceedings of the IEEE/CVF conference on computer vision and pattern recognition},
  pages={17853--17862},
  year={2023}
}

@inproceedings{ved1,
  title={LAVE: LLM-powered agent assistance and language augmentation for video editing},
  author={Wang, Bryan and Li, Yuliang and Lv, Zhaoyang and Xia, Haijun and Xu, Yan and Sodhi, Raj},
  booktitle={Proceedings of the 29th International Conference on Intelligent User Interfaces},
  pages={699--714},
  year={2024}
}

@inproceedings{ved2,
  title={Dragdiffusion: Harnessing diffusion models for interactive point-based image editing},
  author={Shi, Yujun and Xue, Chuhui and Liew, Jun Hao and Pan, Jiachun and Yan, Hanshu and Zhang, Wenqing and Tan, Vincent YF and Bai, Song},
  booktitle={Proceedings of the IEEE/CVF Conference on Computer Vision and Pattern Recognition},
  pages={8839--8849},
  year={2024}
}

@article{luo2025empirical,
  title={An empirical study of catastrophic forgetting in large language models during continual fine-tuning},
  author={Luo, Yun and Yang, Zhen and Meng, Fandong and Li, Yafu and Zhou, Jie and Zhang, Yue},
  journal={IEEE Transactions on Audio, Speech and Language Processing},
  year={2025},
  publisher={IEEE}
}

@misc{liu2025continuallearningvlmssurvey,
      title={Continual Learning for VLMs: A Survey and Taxonomy Beyond Forgetting}, 
      author={Yuyang Liu and Qiuhe Hong and Linlan Huang and Alexandra Gomez-Villa and Dipam Goswami and Xialei Liu and Joost van de Weijer and Yonghong Tian},
      year={2025},
      eprint={2508.04227},
      archivePrefix={arXiv},
      primaryClass={cs.CV},
      url={https://arxiv.org/abs/2508.04227}, 
}

@inproceedings{lin2024video,
  title={Video-llava: Learning united visual representation by alignment before projection},
  author={Lin, Bin and Ye, Yang and Zhu, Bin and Cui, Jiaxi and Ning, Munan and Jin, Peng and Yuan, Li},
  booktitle={Proceedings of the 2024 Conference on Empirical Methods in Natural Language Processing},
  pages={5971--5984},
  year={2024}
}

@article{wang2024qwen2,
  title={Qwen2-vl: Enhancing vision-language model's perception of the world at any resolution},
  author={Wang, Peng and Bai, Shuai and Tan, Sinan and Wang, Shijie and Fan, Zhihao and Bai, Jinze and Chen, Keqin and Liu, Xuejing and Wang, Jialin and Ge, Wenbin and others},
  journal={arXiv preprint arXiv:2409.12191},
  year={2024}
}

@inproceedings{bao2024e3m,
  title={E3m: zero-shot spatio-temporal video grounding with expectation-maximization multimodal modulation},
  author={Bao, Peijun and Shao, Zihao and Yang, Wenhan and Ng, Boon Poh and Kot, Alex C},
  booktitle={European Conference on Computer Vision},
  pages={227--243},
  year={2024},
  organization={Springer}
}

@inproceedings{cheng2024seeclick,
  title={Seeclick: Harnessing gui grounding for advanced visual gui agents},
  author={Cheng, Kanzhi and Sun, Qiushi and Chu, Yougang and Xu, Fangzhi and YanTao, Li and Zhang, Jianbing and Wu, Zhiyong},
  booktitle={Proceedings of the 62nd Annual Meeting of the Association for Computational Linguistics (Volume 1: Long Papers)},
  pages={9313--9332},
  year={2024}
}

@inproceedings{you2024ferret,
  title={Ferret-ui: Grounded mobile ui understanding with multimodal llms},
  author={You, Keen and Zhang, Haotian and Schoop, Eldon and Weers, Floris and Swearngin, Amanda and Nichols, Jeffrey and Yang, Yinfei and Gan, Zhe},
  booktitle={European Conference on Computer Vision},
  pages={240--255},
  year={2024},
  organization={Springer}
}

@misc{houlsby2019parameterefficienttransferlearningnlp,
      title={Parameter-Efficient Transfer Learning for NLP}, 
      author={Neil Houlsby and Andrei Giurgiu and Stanislaw Jastrzebski and Bruna Morrone and Quentin de Laroussilhe and Andrea Gesmundo and Mona Attariyan and Sylvain Gelly},
      year={2019},
      eprint={1902.00751},
      archivePrefix={arXiv},
      primaryClass={cs.LG},
      url={https://arxiv.org/abs/1902.00751}, 
}

@inproceedings{garg2025stpro,
  title={Stpro: Spatial and temporal progressive learning for weakly supervised spatio-temporal grounding},
  author={Garg, Aaryan and Kumar, Akash and Rawat, Yogesh S},
  booktitle={Proceedings of the Computer Vision and Pattern Recognition Conference},
  pages={3384--3394},
  year={2025}
}

@inproceedings{wei2025realvg,
  title={RealVG: Unleashing MLLMs for Training-Free Spatio-Temporal Video Grounding in the Wild},
  author={Wei, Hongchen and Chen, Zhenzhong},
  booktitle={Proceedings of the 33rd ACM International Conference on Multimedia},
  pages={4271--4280},
  year={2025}
}

@inproceedings{zhao2024video,
  title={Video-text prompting for weakly supervised spatio-temporal video grounding},
  author={Zhao, Heng and Yinjie, Zhao and Wen, Bihan and Ong, Yew-Soon and Zhou, Joey Tianyi},
  booktitle={Proceedings of the 2024 Conference on Empirical Methods in Natural Language Processing},
  pages={19494--19505},
  year={2024}
}
